\documentclass[letterpaper, 10 pt, journal, twoside]{IEEEtran}
\usepackage{amsmath,amsfonts,amssymb}
\usepackage{algorithmic}
\usepackage{algorithm}
\usepackage{array}
\usepackage[caption=false,font=normalsize,labelfont=sf,textfont=sf]{subfig}
\usepackage{textcomp}
\usepackage{stfloats}
\usepackage{url}
\usepackage{verbatim}
\usepackage{graphicx}
\usepackage{cite}
\def\BibTeX{{\rm B\kern-.05em{\sc i\kern-.025em b}\kern-.08em
    T\kern-.1667em\lower.7ex\hbox{E}\kern-.125emX}}
\usepackage[pdfstartview=XYZ,
bookmarks=true,
colorlinks=true,
linkcolor=black,
urlcolor=black,
citecolor=black,
bookmarks=true,
linktocpage=true, 
hyperindex=true
]{hyperref}
\usepackage{orcidlink}
\usepackage{balance}

\begin{document}

© 2022 IEEE. Personal use of this material is permitted.
Permission from IEEE must be obtained for all other uses,
including reprinting/republishing this material for advertising
or promotional purposes, collecting new collected works
for resale or redistribution to servers or lists, or reuse of
any copyrighted component of this work in other works.
This work has been submitted to the IEEE for possible
publication. Copyright may be transferred without notice,
after which this version may no longer be accessible.

Please cite the newer, accepted version that has the DOI below:
Digital Object Identiﬁer 10.1109/LRA.2023.3302616
\title{Nonlinear Subsystem-based Adaptive Impedance Control of Physical Human-Robot-Environment Interaction in Contact-rich Tasks}
\author{Mahdi Hejrati \orcidlink{0000-0002-8017-4355}, Jouni Mattila \orcidlink{0000-0003-1799-4323}%
\thanks{ Manuscript received: March 2, 2023; Revised May 10, 2023; Accepted July 22, 2023. This paper was recommended for publication by Editor Jaydev P. Desai upon evaluation of the Associate Editor and Reviewers' comments. The TITAN project is funded by the Technology Industries of Finland Centennial Foundation and the Jane and Aatos Erkko Foundation Future Makers programme. 2020-2023. (Corresponding author: Mahdi Hejrati.)}%
\thanks{ The authors are with the Department of Engineering and Natural Science, Tampere University, 7320 Tampere, Finland {\tt\footnotesize mahdi.hejrati@tuni.fi}; \tt\footnotesize Jouni.Mattila@tuni.fi.}%
\thanks{Digital Object Identifier (DOI): 10.1109/LRA.2023.3302616}
}
\markboth{IEEE Robotics and Automation Letters. Preprint Version. Accepted July 2023}
{Hejrati \MakeLowercase{\textit{et al.}} Nonlinear Subsystem-based Adaptive Impedance Control of pHREI in Contact-rich Tasks} 
\maketitle
\begin{abstract}

Haptic upper limb exoskeletons are robots that assist human operators during task execution while having the ability to render virtual or remote environments. Therefore, ensuring the stability of such robots in physical human-robot-environment interaction (pHREI) is crucial. Having a wide range of Z-width, which indicates the region of passively renderable impedance by a haptic display, is also important for rendering a broad range of virtual environments. To address these issues, this study designs subsystem-based adaptive impedance control to achieve a stable pHREI for 7 degrees of freedom haptic exoskeleton. The presented controller decomposes the entire system into subsystems and designs the controller at the subsystem level. The stability of the controller in the presence of contact with a virtual environment and human arm force is proven by employing the concept of virtual stability. Additionally, the Z-width of the 7-DoF haptic exoskeleton is illustrated using experimental data and improved by exploiting varying virtual mass element. Experimental results are provided to demonstrate the performance of the controller. The control results are also compared to state-of-the-art control methods, highlighting the excellence of the designed controller.
\end{abstract}


\section{Introduction}

\IEEEPARstart{W}{ith} advancements in technology related to the Fourth Industrial Revolution, robots are increasingly being utilized in industries to reduce the workload of workers and in healthcare to improve the quality of human life, showing the significance of human-robot interaction (HRI). HRI can be divided into three categories of human-robot coexistence, human-robot cooperation, and human-robot collaboration or physical HRI (pHRI) \cite{c1}. Unlike to first two categories, in the pHRI scenario, the human and robot are working together on the same task while having a physical impact on each other. An upper limb exoskeleton (ULE) is one example of such an interaction \cite{c2}. On the other hand, a haptic ULE (HULE) \cite{c3} incorporates haptic display technology to render a virtual or remote environment along with the ULE assistance feature. HULE can either be utilized as a master robot in bilateral teleoperation control with the ability to display a remote environment or assist a human during contact-rich co-manipulation (Fig. 1). Consequently, HULE is impacted by human upper limb dynamics along with contact with the virtual or physical environments. Therefore, it is crucial to ensure the stability of physical human-robot-environment interaction (pHREI) in HULE while contact-rich tasks are executed. pHREI control can be implemented by means of the impedance control scheme \cite{c4} that provides the robot with safe and robust compliant behavior throughout interactions with humans and the environment. Additionally, for a haptic display, the Z-width is a key feature that represents the dynamic range of passive impedances that can be rendered. With a larger Z-width, HULE can display a larger range of virtual or remote environments with better feelings \cite{c5}. For this reason, Z-width is an important consideration for haptic display, and maximizing it is desirable.
\begin{figure}[t]
      \centering
      \includegraphics[width=3in]{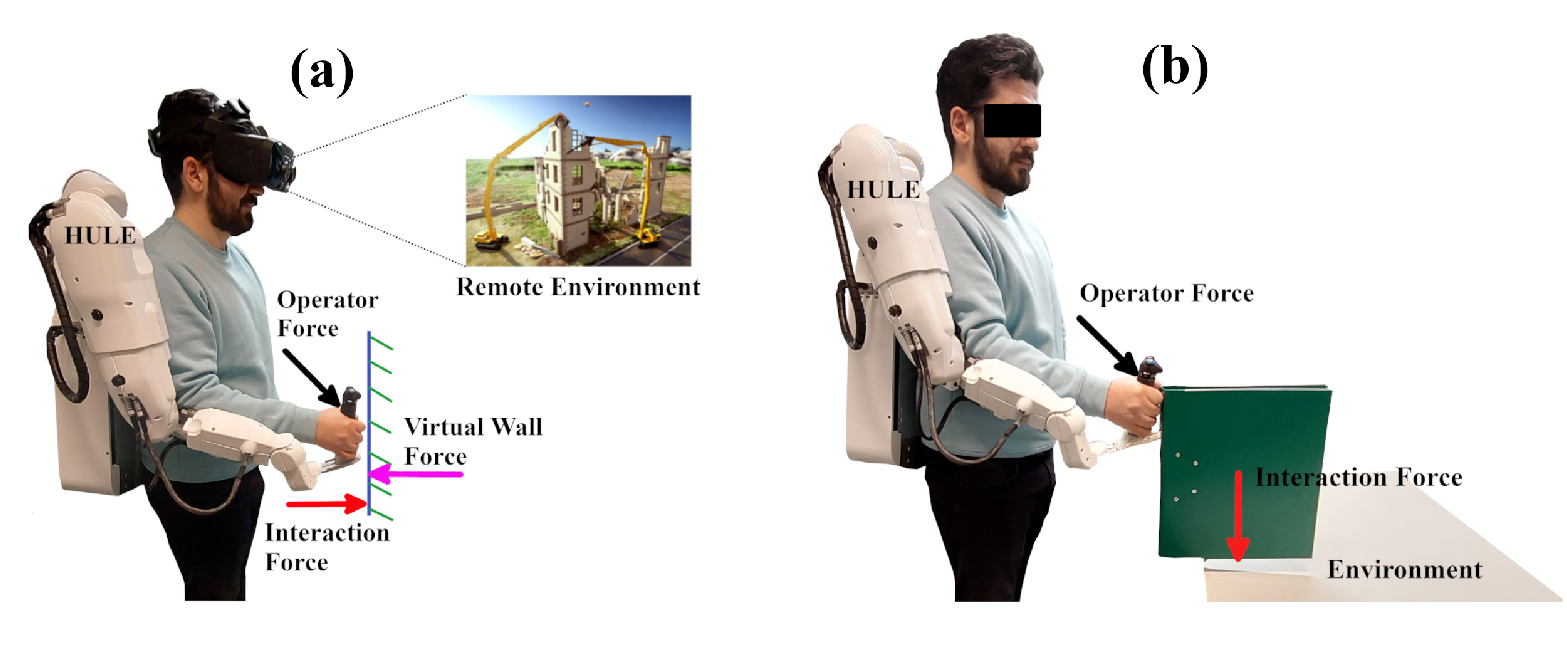}
      \caption{ pHREI for HULE, a) as a master robot in bilateral teleoperation b) as an assisting device for co-manipulation}
      \label{HULEunified}
   \end{figure}
   
\section{Related Works}
\subsection{Impedance Control in pHRI}

Throughout the pHREI, the goal of HULE is to accomplish the task (which can be co-manipulation or assistance) despite the impact of human upper limb dynamics and contact with the environment. Impedance control, which has been widely utilized in the field of robotics\cite{c6,c7}, can equip HULE with such an ability. A multi-point impedance control has been designed to control the pHREI of a 2-DoF robot in \cite{c8}. In \cite{c9}, an adaptive impedance control has been designed to control ULE, in which the impedance parameters of the controller are tuned using surface electromyography and the musculoskeletal model of the human upper limb. In \cite{c10}, an adaptive impedance control has been designed for ULE control based on the muscle circumference sensor for human intention estimation. The goal of that study is to assist a human while carrying an object by using 7-Dof ULE with 5 passive and 2 active DoFs. Adaptive backstepping impedance control is proposed in \cite{c11} for rehabilitation applications using 7-DoF ULE. Moreover, in \cite{c12} iterative learning impedance control has been proposed to control a 1-DoF ULE driven by series elastic actuators, and the desired impedance has been achieved through the iterative manner of the controller. Impedance control has been utilized to control the haptic display as well. In \cite{c13}, nonlinear model reference adaptive impedance control has been proposed to control 5-DoF haptic display. Weighted admittance-impedance control is designed for pHREI control of a 2-DoF haptic device in \cite{c14}. Time Domain Passivity Approach (TDPA) is employed to stabilize the haptic interface by using a passivity controller and passivity observer\cite{c15}. All the mentioned works examined either ULE or haptic devices with low DoFs, whereas, in this study, HULE is a 7-DoF wearable haptic display (Fig. 2(a) shows the HULE in contact with an environment and Fig. 2(b) displays the joint configurations). Therefore, it is crucial to ensure the stability and performance of the controller in the presence of human and environmental impact along with considering the complexity of the system, which guarantees human safety inside HULE.

\subsection{Z-width}
 
Generally, an environment can be represented by a second-order mass-spring-damper impedance model,
\begin{equation}\label{equ1}
    Z(s) = Ms + B + \frac{K}{s}
\end{equation}
with M, B, and K being the mass, damping, and stiffness coefficients of the environment, respectively. As the virtual wall enables us to examine the haptic display and impedance control performance in both high-impedance (having contact with a virtual wall) and low-impedance (free motion) scenarios, in this study, we considered the environment as a unilateral constraint (wall). A larger Z-width means that the haptic display can render a virtual wall with a wider range of stiffness. Therefore, it is important to have a larger Z-width in HULE. Numerous studies have attempted to enhance the Z-width of a haptic display by examining various issues. In \cite{c16,c17}, the effect of an increase in sampling rate has been analyzed. A dual-rate sampling method is proposed in \cite{c18}. However, adding virtual coupling to haptic devices has been a significant approach to increasing the range of the Z-width, as introduced in \cite{c19}. In several works, virtual damping has been proposed to enhance the Z-width \cite{c15,c20}. Nevertheless, in \cite{c21}, it has been shown that using virtual inertia can expand the Z-width of a 1-Dof haptic device more than when virtual damping is employed. All previous works analyzed the Z-width for unwearable low-DoF haptic devices only for performance analysis. By contrast, for high-DoF nonlinear haptic exoskeletons (Fig. 2(a)), it must be ensured that the achieved Z-width is passive in the presence of human effects and extracted for real-world applications.
\subsection{Aims and Contributions}
Based on the literature overview, the aim of the present study is to develop a high-performance HULE with stable interactions with the virtual environment in the presence of human arm dynamics. The contributions of this paper are as follows. First, a subsystem-based adaptive impedance scheme is designed to control the end-effector pose of 7-DoF HULE in the contact-rich task. The proposed controller decomposes the entire system into subsystems, which enables us to design a controller without dealing with the coupled rigid body-actuator nonlinearities at the subsystem level, yet considering it in the overall control law. The natural adaptation law (NAL) is incorporated into the controller to estimate the unknown parameters of the HULE by requiring only one adaptation gain \cite{c22}. NAL also ensures the physical consistency condition for the estimated parameters. The performance of the controller is evaluated by performing experiments and comparing simulation results with other methods. Second, the Z-width of HULE is experimentally drawn in the presence of human arm dynamics and impedance controller by considering all nonlinearities of 7-DoF HULE, which makes the derived achievable Z-width much more reliable in a real-world application. Additionally, varying virtual mass element is employed as a new damping element to enhance the Z-width of HULE.

The paper is organized as follows. In Section III, the mathematical foundation of the approach is described, and the problem is formulated. In Section IV, the proposed subsystem-based impedance control is presented, and in Section V, details of Z-width improvement are given. Experimental results are provided in Section VI with a conclusion part in Section VII.
   
\section{Subsystem-Based Control Scheme}
\subsection{Virtual Decomposition Control}
Virtual decomposition control (VDC) \cite{c23} is a model-based control approach that is suitable for highly nonlinear systems. To design the control action, VDC uses the virtual cutting point (VCP) concept. VCP is a separate interface that essentially cuts through a link. The separated parts formed by the VCP remain in the same position and orientation with applied force/moment with the same magnitude but in opposite directions (Fig. 2(c)). Additionally, VCP distributes the control objective of the entire system to the local control objectives of subsystems with rigid body and actuator parts. Then, based on the Newton-Euler iterative method for a given required velocity, VDC computes the required forces and torques to accomplish each local control goal. Required velocity is the design variable in VDC and must be designed based on the given task. The stability of each subsystem is proved at the local subsystem level and expanded to the stability of the entire complex system by employing a scalar term called the virtual power flow (VPF) and virtual stability scheme. As a result, regardless of how complex a robotic system is, VDC deals with subsystems separately as shown in Fig. 2(c), and designs the controller for each subsystem without dealing with coupled nonlinearities. Thereby, VDC is specifically designed for controlling complex systems, such as pHREI control of a 7-DoF HULE. Recently, VDC scheme has been utilized in various fields, such as teleoperation \cite{c24}, impedance control of hydraulic manipulators \cite{c25}, and ULE control \cite{c22}.

\begin{figure}[t]
      \centering
      \subfloat[]{\includegraphics[width = 0.2\textwidth]{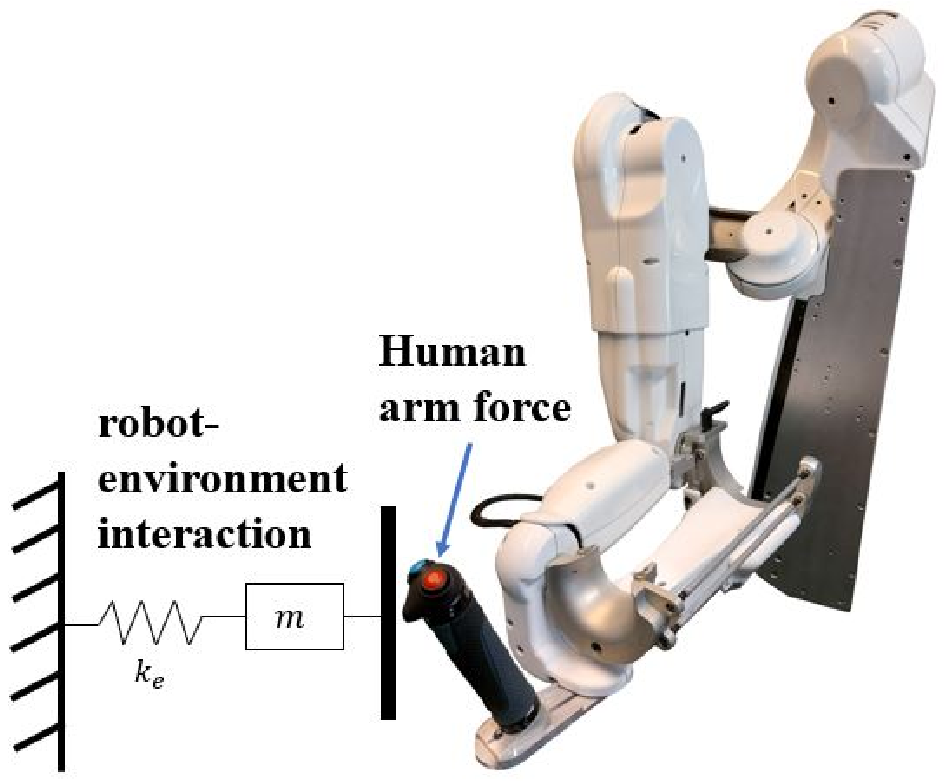}
      \centering
      \label{fig4}}
      \hfil
      \subfloat[]{\includegraphics[width = 0.22\textwidth]{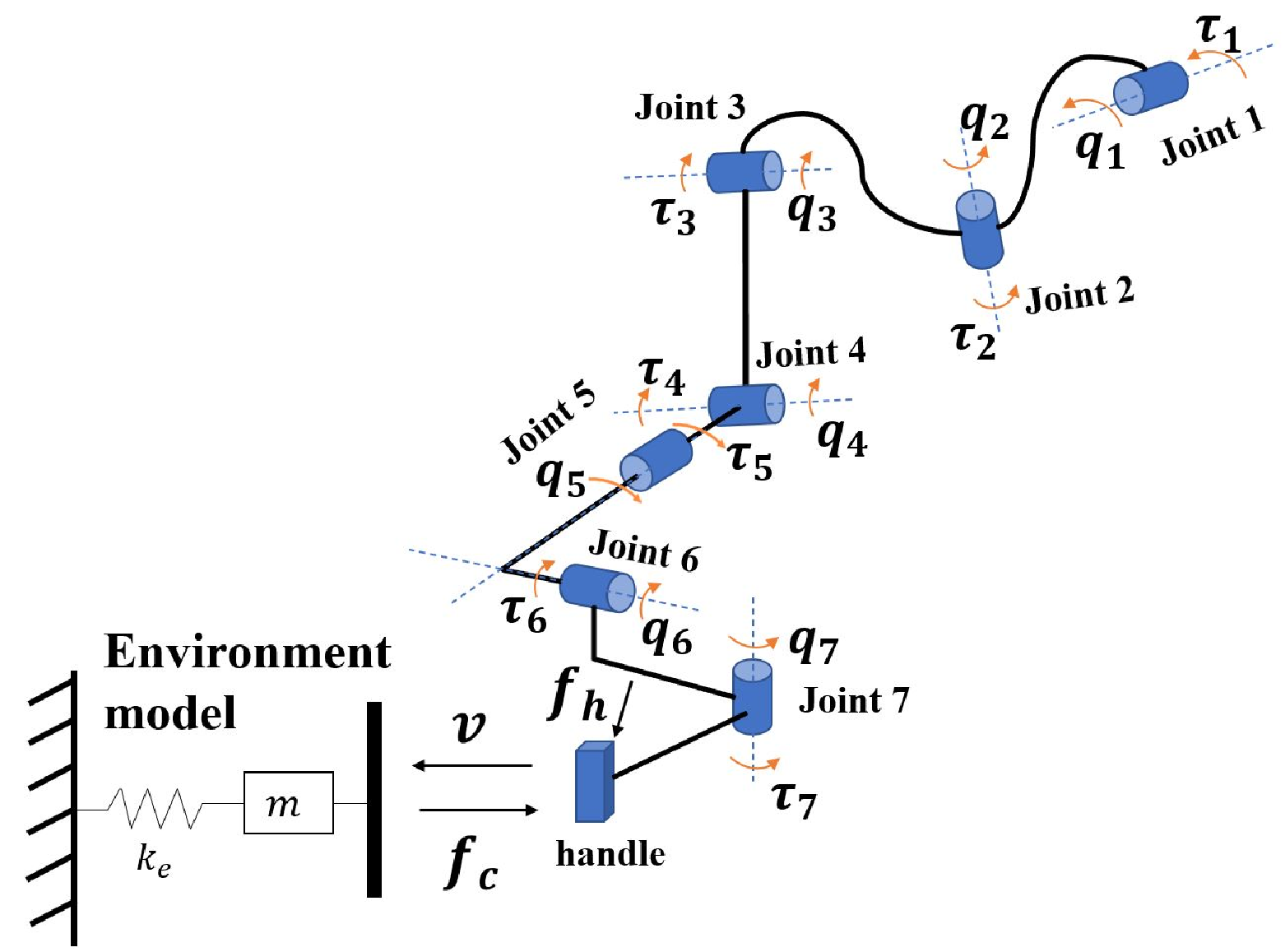}
      \centering
      \label{fig5}}
      \hfil
      \subfloat[]{\includegraphics[scale = 0.25]{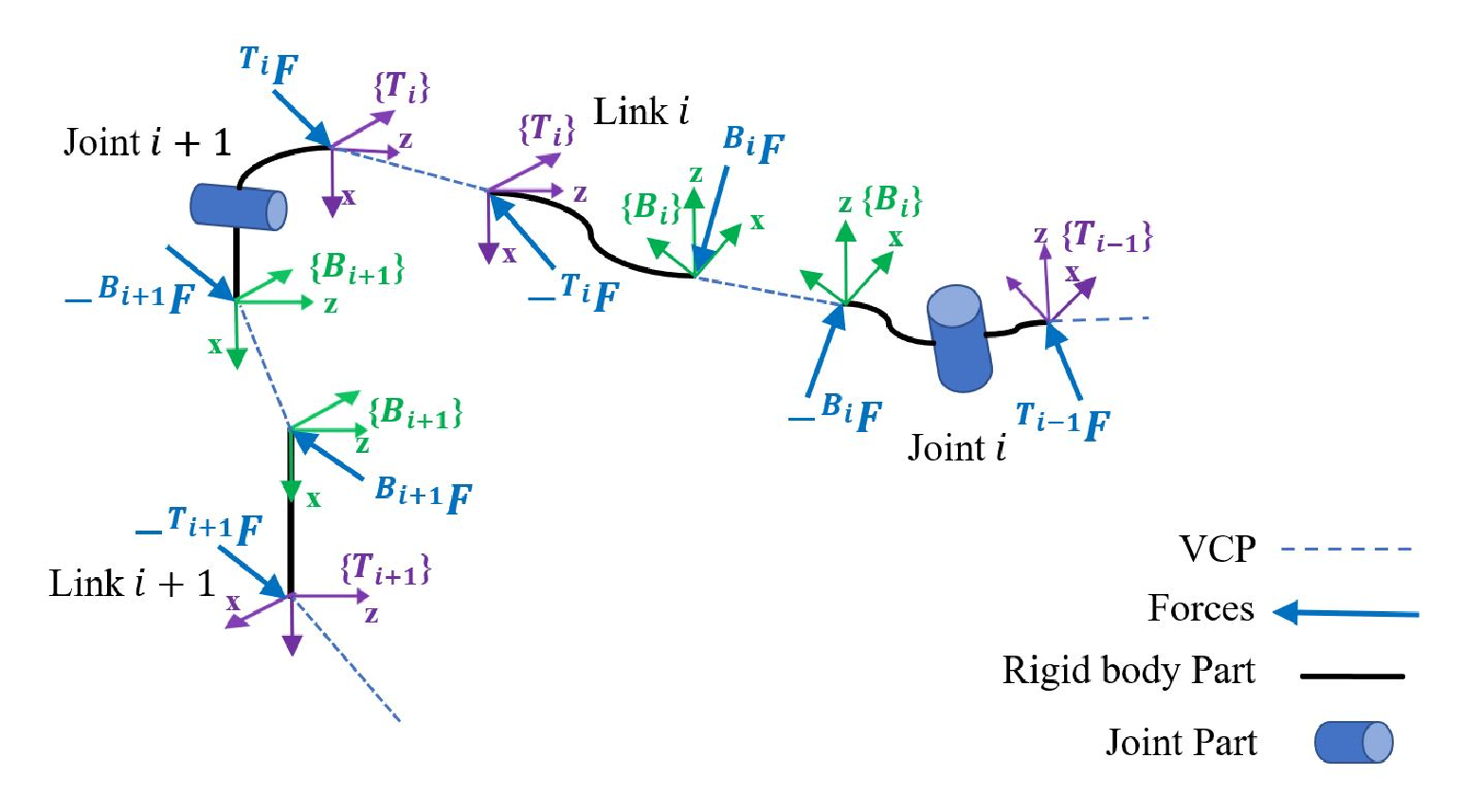}
      \centering
      \label{fig6}}
      \caption{a) HULE in contact with a virtual wall, b) joint configuration, c) demonstration of the decomposition of the HULE for an ith link and actuator}
      \label{HULE confg}
   \end{figure}

\subsection{VDC Mathematical Preliminaries}
\textbf{Deﬁnition 1.} \cite{c23} For a given frame \(\{B_i\}\) at the VCP (Fig. 2(c)), the VPF is defined as,
\begin{equation}\label{equ2}
p_{B_i} = (^{B_i}\mathcal{V}_r-^{B_i}\mathcal{V})^T(^{B_i}F_r-\,^{B_i}F).
\end{equation}

\textbf{Deﬁnition 2.} \cite{c23} A subsystem that is virtually decomposed from a complex robot is said to be virtually stable with its affiliated vectors \(\mathcal{X}(t)\) and \(y(t)\) being a virtual function, if and only if there exists a non-negative accompanying function
\begin{equation}\label{equ3}
\nu(t) \geq \frac{1}{2}\mathcal{X}(t)^TP\mathcal{X}(t)
\end{equation}
such that,
\begin{equation}\label{equ4}
\Dot{\nu}(t) \leq -y(t)^T\,Q\,y(t)+p_A-p_C
\end{equation}
where \(P\) and \(Q\) are two block-diagonal positive-definite matrices, and \(p_A\) and \(p_C\) denote VPFs in the sense of Definition 1 at frames \{A\} (placed at the driven VCPs) and \{C\} (placed at the driving VCPs).

\textbf{Theorem 1.} \cite{c23} Consider a complex robot that is virtually decomposed into subsystems. If all the decomposed
subsystems are virtually stable in the sense of Definition 1, then the entire system is stable.

Theorem 1 is the most important theorem in the VDC context. It establishes the equivalence between virtual stability of every subsystem and stability of the entire complex robot.

\textbf{Lemma 1.} \cite{c22} For any inertial parameter vector \(\phi_{A}\), there is a one-to-one linear map \(f:\Re^{10} \rightarrow S(4)\) such that,
\begin{equation*}
    f(\phi_{A})= \mathcal{L}_{A} = \begin{bmatrix}
        0.5tr(\Bar{I}).\textbf{1}-\Bar{I} & h \\
        h^T & m
        \end{bmatrix}
\end{equation*}
\begin{equation*}
    f^{-1}(\phi_{A}) = \phi_{A}(m,h,tr(\Sigma).\textbf{1}-\Sigma)
\end{equation*}
where \(\Sigma = 0.5tr(\Bar{I})-\Bar{I}\), and \(m\),\(h\), and \(\Bar{I}\) are the mass, first mass moment, and rotational inertia matrix, respectively, with \(\phi_{A} = [m,h,vecI]\) where \(vecI = [\Bar{I}_{xx},\Bar{I}_{yy},\Bar{I}_{zz},\Bar{I}_{xy},\Bar{I}_{yz},\Bar{I}_{xz}]\). Moreover, \(tr(.)\) is the Trace operator of a matrix. Then, the NAL adaptation law can be written as,
\begin{equation*}
    \Dot{\hat{\mathcal{L}}}_{A} = \frac{1}{\gamma}\hat{\mathcal{L}}_{A}\,\mathcal{S}_{A}\,\hat{\mathcal{L}}_{A}
\end{equation*}
where \(\hat{\mathcal{L}}_{A}\) is estimation of \(\mathcal{L}_{A}\), \(\gamma\) is the adaptation gain, and \(\mathcal{S}_{A}\) is a unique symmetric matrix.

\textbf{Lemma 2.} \cite{c22} For \(\mathcal{L}_{A}\) defined in Lemma 1, Bregman divergence with the log-det function can be defined as,
\begin{equation*}
    \mathcal{D}_F(\mathcal{L}_{A}\rVert \hat{\mathcal{L}}_{A}) = log\frac{|\hat{\mathcal{L}}_{A}|}{|\mathcal{L}_{A}|}+tr(\hat{\mathcal{L}}_{A}^{-1}\mathcal{L}_{A})-4.
\end{equation*}
with the time derivative of,
\begin{equation*}
    \Dot{\mathcal{D}}_F(\mathcal{L}_{A}\rVert \hat{\mathcal{L}}_{A}) = tr([\hat{\mathcal{L}}_{A}^{-1}\Dot{\hat{\mathcal{L}}}_{A}\,\hat{\mathcal{L}}_{A}^{-1}]\,\Tilde{\mathcal{L}}_{A})
\end{equation*}
where \(\Tilde{\mathcal{L}}_{A} = \hat{\mathcal{L}}_{A} - \mathcal{L}_{A}\). 

\subsection{VDC Control Design}

Consider \(\{B_i\}\) and \(\{T_i\}\) as frames that are attached to the \(i^{th}\) rigid body. Then, the 6D linear/angular velocity vector \(^{B_i}\mathcal{V}\in \Re^6\) and force/moment vector \(^{B_i}\mathcal{F}\in \Re^6\) can be expressed as follows \cite{c23}:
\begin{equation*}
^{B_i}\mathcal{V} = [\,^{B_i}v,\,^{B_i}\omega]^T,\quad ^{B_i}\mathcal{F} = [\,^{B_i}f,\,^{B_i}m]^T
\end{equation*}
where \(^{B_i}v\in \Re^3\) and \(^{B_i}\omega\in \Re^3\) are the linear and angular velocities of frame \(\{B_i\}\), and \(^{B_i}f\in \Re^3\) and \(^{B_i}m\in \Re^3\) are the force and moment expressed in frame \(\{B_i\}\), respectively. The transformation matrix that transforms force/moment vectors and velocity vectors between frames \(\{B_i\}\) and \(\{T_i\}\) is \cite{c23},
\begin{equation}\label{equ5}
^{B_i}U_{T_i} = \begin{bmatrix}
^{B_i}R_{T_i} & \textbf{0}_{3\times3} \\
(^{B_i}r_{{B_i}{T_i}}\times)\, ^{B_i}R_{T_i} & ^{B_i}R_{T_i}
\end{bmatrix}
\end{equation}
where \(^{B_i}R_{T_i} \in \Re^{3\times3} \) is a rotation matrix between frame \(\{B_i\}\) and \(\{T_i\}\), and (\(^{B_i}r_{{B_i}{T_i}}\times\)) is a skew-symmetric matrix operator defined in \cite{c23}. Based on \(^{B_i}U_{T_i}\), force/moment and velocity vectors can be transformed between frames as \cite{c23},
\begin{equation}\label{equ6}
^{T_i}\mathcal{V} =\, ^{B_i}U_{T_i}^T\,^{B_i}\mathcal{V}, \quad i = 1... 7
\end{equation}
\begin{equation}\label{equ7}
^{B_i}\mathcal{V} =\, \kappa_i \Dot{q}_{i} +\, ^{T_{i-1}}U_{B_i}^T\,^{T_{i-1}}\mathcal{V}, \quad i = 1... 7
\end{equation}
\begin{equation}\label{equ8}
^{B_i}F =\, ^{B_i}F^* +\, ^{B_i}U_{T_i}\, ^{T_i}F, \quad i = 7... 1
\end{equation}
with \(^{T_{i-1}}F =\, ^{T_{i-1}}U_{B_i}\, ^{B_i}F\) for \(i = 7... 1\), \(q_i\) and \(\Dot{q}_i\) being joint angles and velocities, respectively, \(\kappa_i = z_\tau\) for \(i = 1,2,3,4,6\), \(\kappa_i = y_\tau\) for \(i = 7\), and \(\kappa_i = x_\tau\) for \(i = 5\), which \(z_\tau = [0,0,0,0,0,1]^T\), \(y_\tau = [0,0,0,0,1,0]^T\), \(x_\tau = [0,0,0,1,0,0]^T\), and \((.)^T\) is Transpose operator. Moreover, for \(i=1\) we have \(^{T_{0}}\mathcal{V} =\, ^{G}\mathcal{V}=\textbf{0}_{6\times1}\) and \(^{T_{0}}F =\textbf{0}_{6\times1}\) , and for \(i=7\), we have \(^{T_{7}}F =f\), where \(f\) is the interaction force and will be defined later. \(^{B_i}F^*\) is the net force applied to the \(i^{th}\) rigid body, defined as,
\begin{equation}\label{equ9}
M_{B_i}\frac{d}{dt}(^{B_i}\mathcal{V})+C_{B_i}(^{B_i}\mathcal{V})+G_{B_i}=\, ^{B_i}F^*
\end{equation}
where \(M_{B_i} \in \Re^{6\times6} \), \(C_{B_i} \in \Re^{6\times6} \), and \(G_{B_i} \in \Re^6 \) are the mass, centrifugal and Coriolis, and gravity matrices defined in \cite{c22}, respectively. The local control objective for the rigid body part is to reach the given required linear/angular velocity \(^{B_i}\mathcal{V}_r\), which will be defined later. Therefore, the required force/moment vector that must be applied to the \(i^{th}\) rigid body to accomplish local control objectives, can be designed as,
\begin{equation}\label{equ10}
^{B_i}F_r =\, ^{B_i}F^*_r +\, ^{B_i}U_{T_i}\,^{T_i}F_r.
\end{equation}
with the local control law of,
\begin{equation}\label{equ11}
^{B_i}F_r^* =\, ^{B_i}W\hat{\phi}_{B_i} +\,K_{Di}\,^{B_i}e_{\mathcal{V}} + \,K_{Ii}\,\int_{0}^{t} \,^{B_i}e_{\mathcal{V}} dt
\end{equation}
where \(^{B_i}e_{\mathcal{V}} = {^{B_i}\mathcal{V}_{r}} - {^{B_i}\mathcal{V}}\), and \(K_{Di}\) and \(K_{Ii}\) are diagonal positive-definite control gains, \(^{B_i}W \in \Re^{6\times10}\) is the regression matrix, and \(\hat{\phi}_{B_i} \in \Re^{10}\) is the estimation of inertial parameter vector \(\phi_{B_i} \in \Re^{10}\) \cite{c22} defined as,
\begin{equation}\label{equ12}
    ^{B_i}W\phi_{B_i} = M_{B_i}\frac{d}{dt}(^{B_i}\mathcal{V}_r)+C_{B_i}(^{B_i}\mathcal{V}_r)+G_{B_i} 
\end{equation}

The joint actuator dynamics for an electrical motor in the absence of friction can be expressed as,
\begin{equation}\label{equ13}
I_{mi} \Ddot{q}_i = \tau^*_i=\tau_i-\tau_{ai}
\end{equation}
where \(I_{mi}\) denotes the moment of inertia of the joint. The equation (\ref{equ13}) is exactly in the same sense as (\ref{equ10}), with \(\tau^*_i\) being the net torque applied to the joint, \(\tau_i\) representing the control torque, and \(\tau_{ai} = z_{\tau}^T\,^{B_i}F\). The adaptive VDC control law for accomplishing both local control objectives can be designed as, 
\begin{equation}\label{equ14}
\tau_i= \tau^*_{ir}+\tau_{air}
\end{equation}
with \(\tau^*_{ir}\) being the local control law that ensures tracking of the desired angle for the joint, defined as,
\begin{equation}\label{equ15}
\tau^*_{ir} = W_{ai}\hat{\phi}_{ai} + k_{di} e_{ai} + k_{Ii} \int_{0}^{t} e_{ai}
\end{equation}
where \(W_{ai} \in \Re^{1\times10}\) is the regression matrix, and \(\hat{\phi}_{ai} \in \Re^{10}\) is the estimation of inertial parameter vector \(\phi_{ai} \in \Re^{10}\) in,
\begin{equation}\label{equ16}
    W_{ai}\phi_{ai} = I_{mi} \Ddot{q}_{ri}.
\end{equation}
Also, \(e_{ai} = \Dot{q}_{ir}-\Dot{q}_i\), \(\Dot{q}_{ir}\) is the required angular velocity, \(k_{di}\) and \(k_{Ii}\) are positive control gains. Moreover, \(\tau_{air} = z_{\tau}^T\, ^{B_i}F_r\) stands for accomplishing the local control objectives of a rigid body part with \(^{B_i}F_r\) being computed from (\ref{equ10}). The estimation of inertial parameter vectors \(\hat{\phi}_{B_i}\) and \(\hat{\phi}_{ai}\) in (\ref{equ11}) and (\ref{equ15}) can be computed by exploiting NAL in the sense of Lemma 1 by replacing \{A\} with \(\{B_i\}\) and \(\{a_i\}\). The adaptive law in Lemma 1 is only utilized for estimation of inertial parameters of the robot and human arm model in (\ref{equ11}) and (\ref{equ15}). As can be seen from Lemma 1, only one gain is required to be tuned for estimation of unknown parameters. Moreover, the only condition to satisfy physical consistency condition of estimated parameters is that \(\hat{\mathcal{L}}_{A}(0)\) be physically consistent \cite{c22}.

The required linear/angular velocity, 
\(^{B_i}\mathcal{V}_r\), can be computed as,
\begin{equation}\label{equ17}
^{B_i}\mathcal{V}_r =\, \kappa_i \Dot{q}_{ri} +\, ^{T_{i-1}}U_{B_i}^T\,^{T_{i-1}}\mathcal{V}_r, \quad i = 1...7
\end{equation}
where \(^{T_{i-1}}\mathcal{V}_r\) is the required velocity of the VCP on the \((i-1)^{th}\) rigid body (for \(i=1\), we have, \(^{T_{0}}\mathcal{V}_r = \,^{G}\mathcal{V}_r = \textbf{0}_{6\times1}\)), and \(\Dot{q}_{ri}\) is the design variable and should be designed based on the predefined task. According to (\ref{equ7}) and (\ref{equ17}), if \(\Dot{q}_i \rightarrow \Dot{q}_{ir}\), then \(^{B_i}\mathcal{V} \rightarrow\, ^{B_i}\mathcal{V}_r\), satisfying both local control objectives. The control law in (\ref{equ14}) is designed in such a way to accomplish this goal. Based on what is explained so far, the most important part of deriving VDC control law (\ref{equ14}) is designing \(\Dot{q}_{ir}\). In this paper, in which the task space problem of a 7-DoF HULE is of concern, \(\Dot{q}_{ri}\) can be defined as,
\begin{equation}\label{equ18}
\Dot{q}_{ri} = J^{T}(JJ^{T})^{-1}\Dot{X}_r
\end{equation}
with \(\Dot{X}_r\) being the task space design variable and \(J \in \Re^{6\times7}\) is the Jacobian matrix. In the next section, \(\Dot{X}_r\) is designed in order to achieve the impedance control goal.

\section{Impedance Control Design}

Impedance control equips the robot with the ability to accomplish both free-motion and contact-rich tasks. By neglecting the inertia in \cite{c4}, the desired impedance model for the robot can be expressed as,
\begin{equation}\label{equ20}
    -B_d(\Dot{X}_d-\Dot{X})-K_d(X_d-X) = f_d-f
\end{equation}
where \(B_d \in \Re^{6\times6}\) and \(K_d \in \Re^{6\times6}\) are diagonal positive-definite matrices and characterize the desired damping and stiffness, respectively. \(f\) is the force applied to the HULE from the outside, \(f_d\) is the desired contact force, and \(X_d\) and \(X\) are desired and actual position of the end-effector, respectively. Moreover, \(e = X_d-X = [e_p, e_o]\), where \(e_p\) is the position error and \(e_o\) is orientation error using the quaternion \cite{c22} of the end-effector. As a haptic exoskeleton, HULE is impacted by human arm force \(f_h\) during the execution of the task and virtual environment force \(f_c\) whenever contact happens. Therefore, the external force applied to the robot during pHREI is different and can be expressed as,
\begin{equation}\label{equ21}
\begin{split}
    f = \begin{cases}
        -f_h & :\,assisting (\textit{pHRI})\\
        -f_h+f_c & :\,contact (\textit{pHREI}).
    \end{cases}
\end{split}
\end{equation}
The dynamic equation governing human arm dynamics can be expressed as,
\begin{equation}\label{equ22}
     M_h \Ddot{X}_h+B_h \Dot{X}_h + K_h (X_h-X) = f_h
\end{equation}
where \(M_h \in \Re^{6\times6}\), \(B_h \in \Re^{6\times6}\), and \(K_h \in \Re^{6\times6}\), are diagonal positive-definite matrices and characterize the inertia, damping, and stiffness, matrices of the human arm, respectively, and \(X_h\) is human hand position. It must be mentioned that human exogenous force is omitted. The \(f_c\) will be defined later. 

As mentioned previously, the required task space velocity in (\ref{equ18}) is the control design variable for accomplishing desired objectives. The following theorem summarizes the subsystem-based impedance control law.

\textbf{Theorem 2.} For a decomposed system represented in (\ref{equ9}) and (\ref{equ13}) with the interaction force of (\ref{equ21}), the desired impedance model (\ref{equ20}) can be achieved for the robot if the design variable in (\ref{equ18}), defined as,
\begin{equation}\label{equ23}
    \Dot{X}_r = \Dot{X}_d + \Gamma_x(X_d-X)+\Gamma_f(f_d-f)
\end{equation}
with \(\Gamma_x \in \Re^{6\times6}\), and \(\Gamma_f \in \Re^{6\times6}\) being positive-definite diagonal matrices defined as,
\begin{equation}\label{equ24}
    \Gamma_f = B_d^{-1}, \qquad \Gamma_x = K_d\,B_d^{-1}.
\end{equation}
It must be mentioned that \(B_d\) and \(K_d\) should be selected from the achievable dynamic range of robot.

\textbf{Proof.} Appendix A.

The following theorem ensures the stability of the entire system in the sense of Theorem 1 and Definition 2. 

\textbf{Theorem 3.} For a decomposed 7-DoF HULE (Fig. 2(c)) with a rigid body and actuator dynamics of (\ref{equ9}) and (\ref{equ13}), control action of (\ref{equ14}), adaptation laws in the sense of Lemma 1, and \(\Dot{X}_r\) defined in (\ref{equ23}), the non-negative accompanying function can be defined as,
\begin{equation}\label{equ25}
\begin{split}
    \nu_i(t) & = \sum_{i=1}^{7} [\frac{1}{2}(\int_{0}^{t}\, ^{B_i}e_{\mathcal{V}} dt)^T\,K_{Ii}\,(\int_{0}^{t}\, ^{B_i}e_{\mathcal{V}} dt)\\
    & +\frac{1}{2}k_{Ii} (\int_{0}^{t} e_{ai} dt)^2 + \,\frac{1}{2}\,^{B_i}e_{\mathcal{V}}^T\,M_{B_i}\,^{B_i}e_{\mathcal{V}} + \, \frac{1}{2}I_{mi}\,e_{ai}^2 \\ 
    & + \gamma\mathcal{D}_F(\mathcal{L}_{B_i}\rVert \hat{\mathcal{L}}_{B_i}) + \mathcal{D}_F(\mathcal{L}_{ai}\rVert \hat{\mathcal{L}}_{ai})]
\end{split}
\end{equation}
where \(\mathcal{D}_F(\mathcal{L}\rVert \hat{\mathcal{L}}) \) is defined in Lemma 2. Taken the time derivative of (\ref{equ25}), we can obtain,
\begin{equation}\label{equ26}
    \Dot{\nu}_i (t) = \sum_{i=1}^{7} (-\,^{B_i}e_{\mathcal{V}}^T\,K_{Di}\,^{B_i}e_{\mathcal{V}}-k_{di}e_{ai}^2).
\end{equation}
where, according to Theorem 1 and Definition 2, the asymptotic stability of the entire system is proven.

\textbf{Proof.} Appendix B.

\textbf{Remark 1.} The designed control law in (\ref{equ14}) has two terms. The second term on the right-hand side of (\ref{equ14}) compensates for the rigid body forces resulting from motion and gravity. The first term, on the other hand, concentrates on achieving the required velocity defined in (\ref{equ18}), which ensures that the desired impedance model defined in (\ref{equ20}) is achieved by the robot. Therefore, the desired impedance is achieved by accomplishing the local objectives.

\section{Z-width of HULE}

For a haptic display, the Z-width is an indicator of the range of stiffness of a virtual wall that can be passively rendered. The passivity condition for a contact with the generated force of \(f_c(t)\) and haptic end-effector velocity of \(\textit{v}(t)\) is defined as,
\begin{equation}\label{equ27}
    E_c(t) = \int_{0}^{t} f_c(\sigma)\,\textit{v}(\sigma)\, d\sigma \geq 0. \qquad \forall t\geq 0
\end{equation}
For a stiff virtual wall, the spring force can be computed as,
\begin{equation}\label{equ28}
    f_s(t) = k_e(z_e-z)
\end{equation}
where \(k_e\) is the rendered stiffness of the virtual wall, \(z_e\) is the position of virtual wall, and \(z\) is the position of HULE end-effector. As can be seen from (\ref{equ28}), increasing the stiffness of the virtual wall generates larger contact forces, resulting in unstable contact. Therefore, it limits the region of the Z-width. One alternative to having a wider Z-width (being able to render a larger \(k_e\) passively) is to add energy dissipative terms to the virtual wall. To do so, we employed the varying virtual mass element. After a contact between HULE end-effector and the virtual environment, unlike virtual damping, varying virtual mass is activated whenever the absolute amount of kinetic energy in the environment increases. Then, it dissipates the amount of energy that is equivalent to acceleration. Based on the explanation, varying virtual mass can be defined as,
\begin{equation}\label{equ29}
    m_e = \begin{cases}
    m_d  & if\, \textit{a}.\textit{v} \geq 0, \\
    0 & else
    \end{cases}
\end{equation}
where \(m_e\) is the virtual mass felt in the contact direction, \(m_d\) is the desired virtual mass that must be displayed, and
\(\textit{a}\) and \(\textit{v}\) are the acceleration and velocity of the HULE end-effector in the direction of contact. The damping force of virtual elements can be computed as,
\begin{equation}\label{equ31}
    f_{damp} = \begin{cases}
    m_e\textit{a}  & \text{for} \quad n= 1 \\
    b_e\textit{v} & \text{for} \quad n= 2
    \end{cases}
\end{equation}
with \(b_e\) being virtual damping coefficient. Finally, the contact force can be computed as,
\begin{equation}\label{equ32}
    f_c(t) = f_s(t) + f_{damp}(t).
\end{equation}
By substituting (\ref{equ31}) and (\ref{equ32}) in (\ref{equ27}), we can obtain,
\begin{equation}\label{equ33}
    E_c(t) = \begin{cases}
    \int_{0}^{t} (k_e(z_e-z)+m_e\textit{a})\,\textit{v}(\sigma)\, d\sigma \geq 0  & \text{for} \quad n= 1 \\
    \int_{0}^{t} (k_e(z_e-z)+b_e\textit{v})\,\textit{v}(\sigma)\, d\sigma \geq 0 & \text{for} \quad n= 2
    \end{cases}
\end{equation}
where \(n = 1\) and \(n = 2\) represents varying virtual mass and virtual damping, respectively. For a given damping element type and velocity and acceleration of the end-effector at the time of contact, the region of passively renderable stiffness can be derived by solving of (\ref{equ33}). However, in this paper, the passive region is extracted experimentally to avoid solving of (\ref{equ33}). It also must be mentioned that for a 1-DoF haptic interface a compact mathematical expression of the (\ref{equ33}) with virtual damping is expressed in \cite{c26}, whereas there is no compact expression for 7-DoF HULE because of the extremely nonlinear and coupled dynamics. In the results section, the generated energy with both designed varying virtual mass and virtual damping is experimentally demonstrated using (\ref{equ33}), and their effect on contact stability is examined. It is shown that employing varying virtual mass in (\ref{equ29}) helps to achieve a larger Z-width.

\begin{figure}[t]
    \centering
    \includegraphics[width=2.2in]{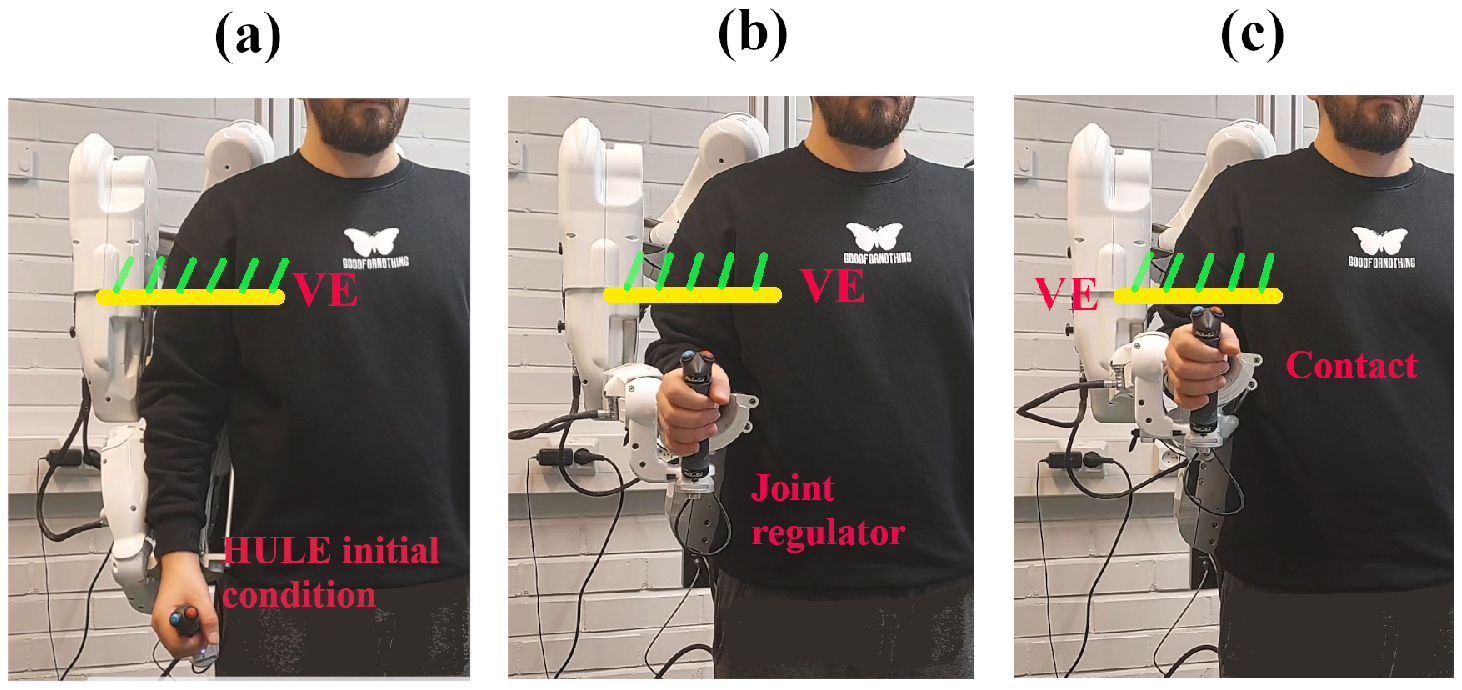}
    \caption{Experiment steps for Z-width plot, a) initial configuration for HULE, b) VDC regulator calibrates joint angles, c) impedance control moves robot and contact happens. (VE: Virtual Environment)}
    \label{figurelabel1}
\end{figure}

\begin{figure}[ht]
    \centering
      \subfloat[]{\includegraphics[width = 0.9\columnwidth]{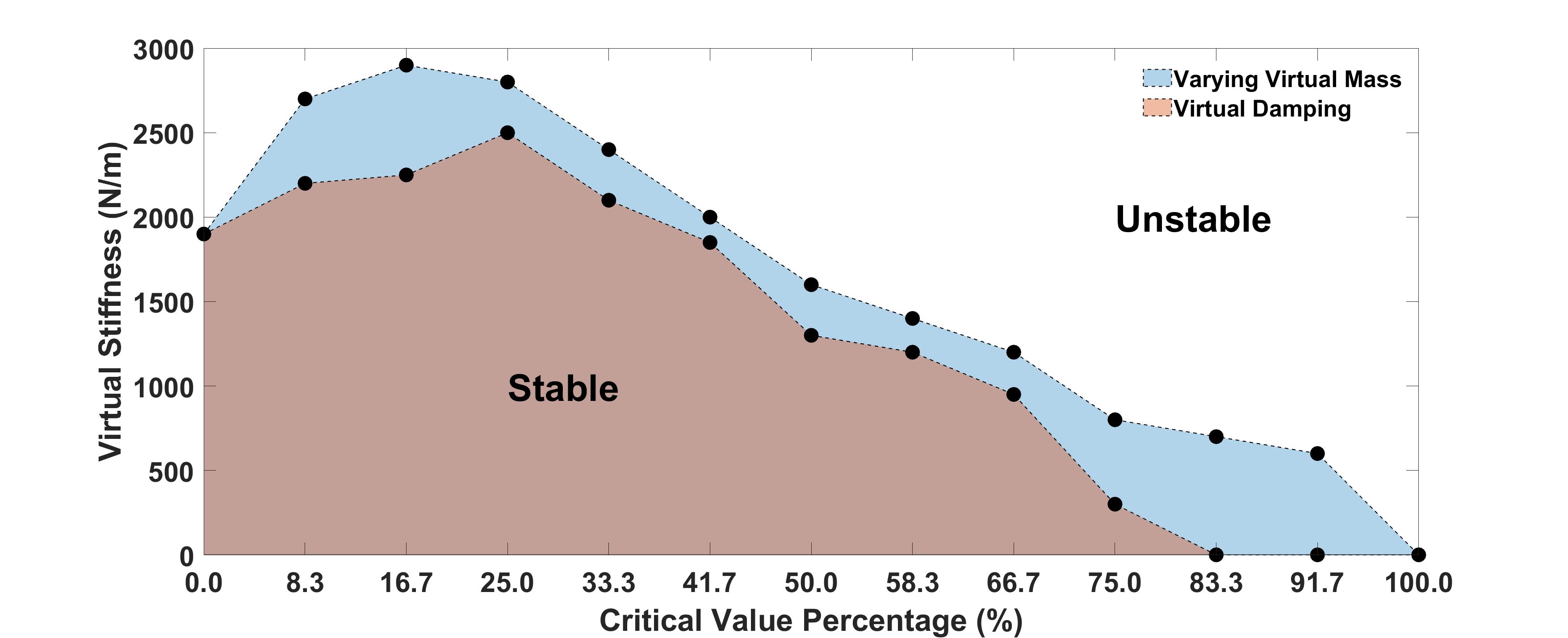}
      \centering
      \label{figg4}}
      \hfil
      \subfloat[]{\includegraphics[width = 0.24\textwidth]{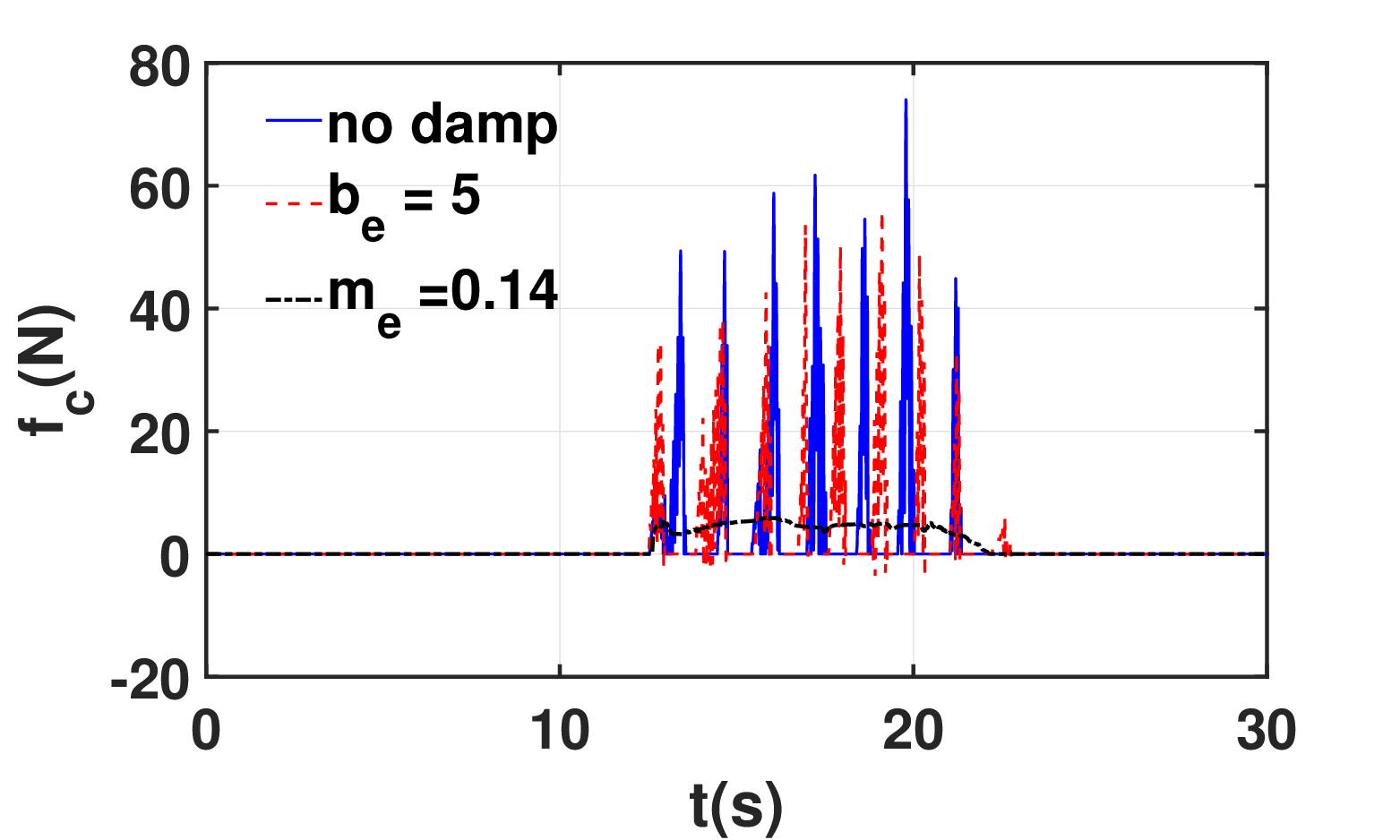}
      \centering
      \label{fig5a}}
      \hfil
      \subfloat[]{\includegraphics[width = 0.22\textwidth]{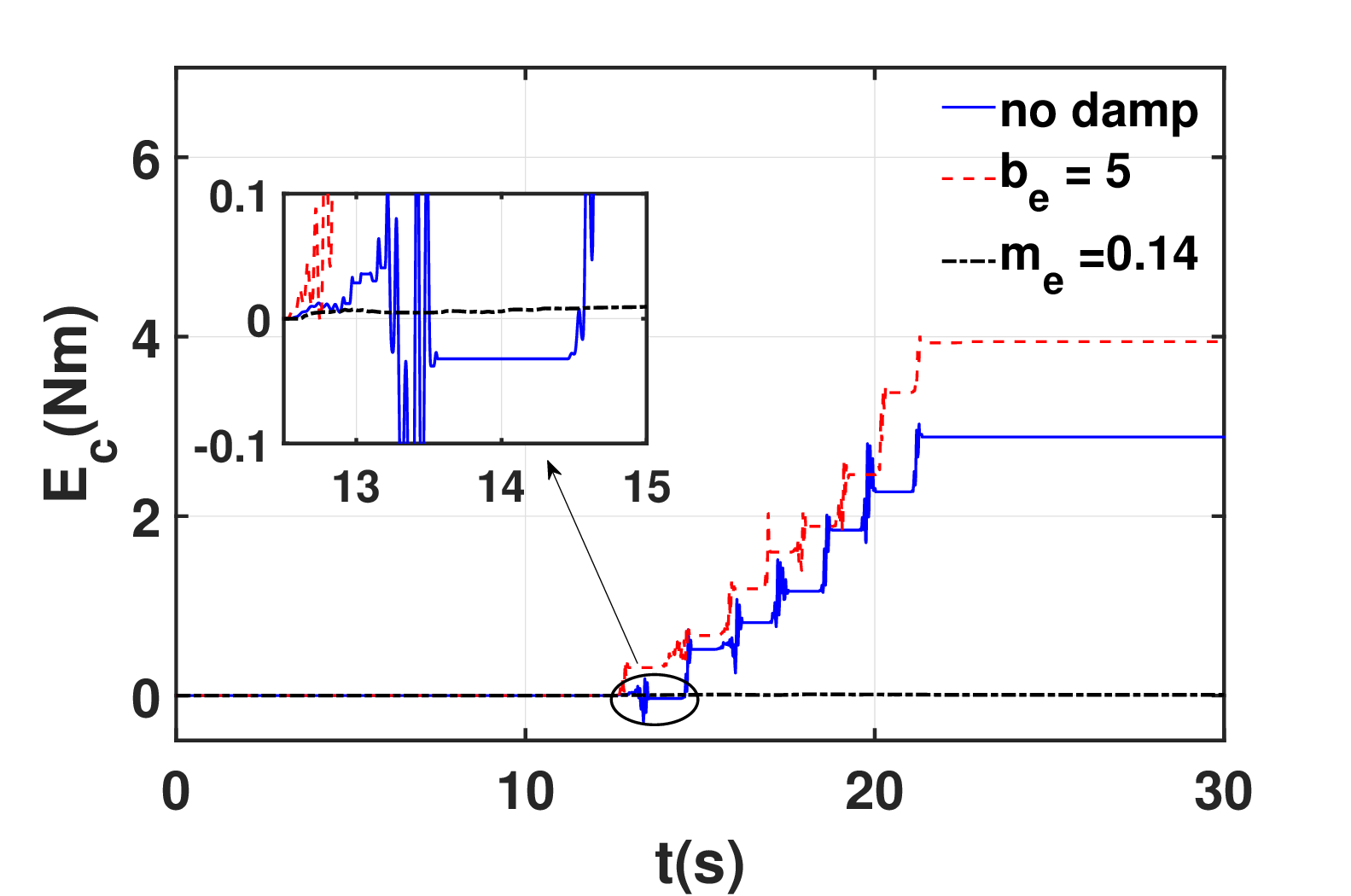}
      \centering
      \label{fig5d}}
      \caption{ (a) Z-width plot of HULE for \(K_d = 100\, N/m\) and \( B_d = 40 \, Ns/m\) with \(t_f = 5\, s\). The light blue area is Z-width with varying virtual mass and the light brown area is with virtual damping. The region under the lines is a passive region. (b) contact force plots and, c) energy generation figure during contact with and without damping elements for the environment with \(k_e = 2500\,N/m\) and \(t_f = 5\, s\).}
    \label{figurelabel2}
\end{figure}

\section{Results and Discussion}
\subsection{Z-width Results}

In this section, the Z-width of the HULE using both virtual damping and varying virtual mass is drawn. Fig. 3 demonstrates the experimental setup and steps for drawing Z-width. When the operator grasps the handle (Fig. 3(a)), HULE is activated at a random initial condition for joint angles. For a better comparison with having the same initial start point, a VDC-based joint controller is employed (Fig. 3(b)) to regulate the joint angles. Then, a \(5^{th}\) order smooth-acceleration trajectory is generated in Cartesian space with the specified time \(t_f\) \cite{c27}. The lower \(t_f\), the faster the trajectory and vice versa. Finally, the proposed impedance control makes HULE follow the desired path (Fig. 3(c)). The desired impedance of the robot in (\ref{equ20}) is fixed to \(K_d = 100\, diag([2,2,1,1,1,1])\, N/m\) and \( B_d = 40\, diag([1,1,1,1,1,1])\, Ns/m\).

For virtual damping (\(b_e\)) varying from 0 to 60 \(Ns/m\) with an interval step of 5 \(Ns/m\), and for virtual mass (\(m_d\)) varying from 0 to 1.68 \(kg\) with an interval of 0.14 \(kg\), the maximum virtual stiffness that can be rendered passively is illustrated in Fig. 4(a). The critical value shows the value for each damping element that results in the rendering of zero stiffness. Using this term gives a better intuition that by changing each damping element from zero to maximum appliable value, how much stiffness can be rendered. As can be seen from Fig. 4(a), employing varying virtual mass expands the Z-width of HULE. Moreover, to demonstrate the effect of adding varying virtual mass, in Fig. 4(b) and 4(c), the result of contact with a virtual wall with \(k_e = 2500\, N/m\) is displayed. Fig. 4(b) and 4(c) demonstrate that the contact with no damping element and with virtual damping \(b_e = 5\,  Ns/m\) is unstable, while with varying virtual mass \(m_d = 0.14\, kg\) passivity condition of (\ref{equ33}) is ensured. Considering all the aforementioned along with uncertainties and nonlinearities, Z-width drawn in Fig. 4(a) is much more reliable and applicable for real-world application. 

\begin{figure}[t]
    \centering
    \includegraphics[width=3in]{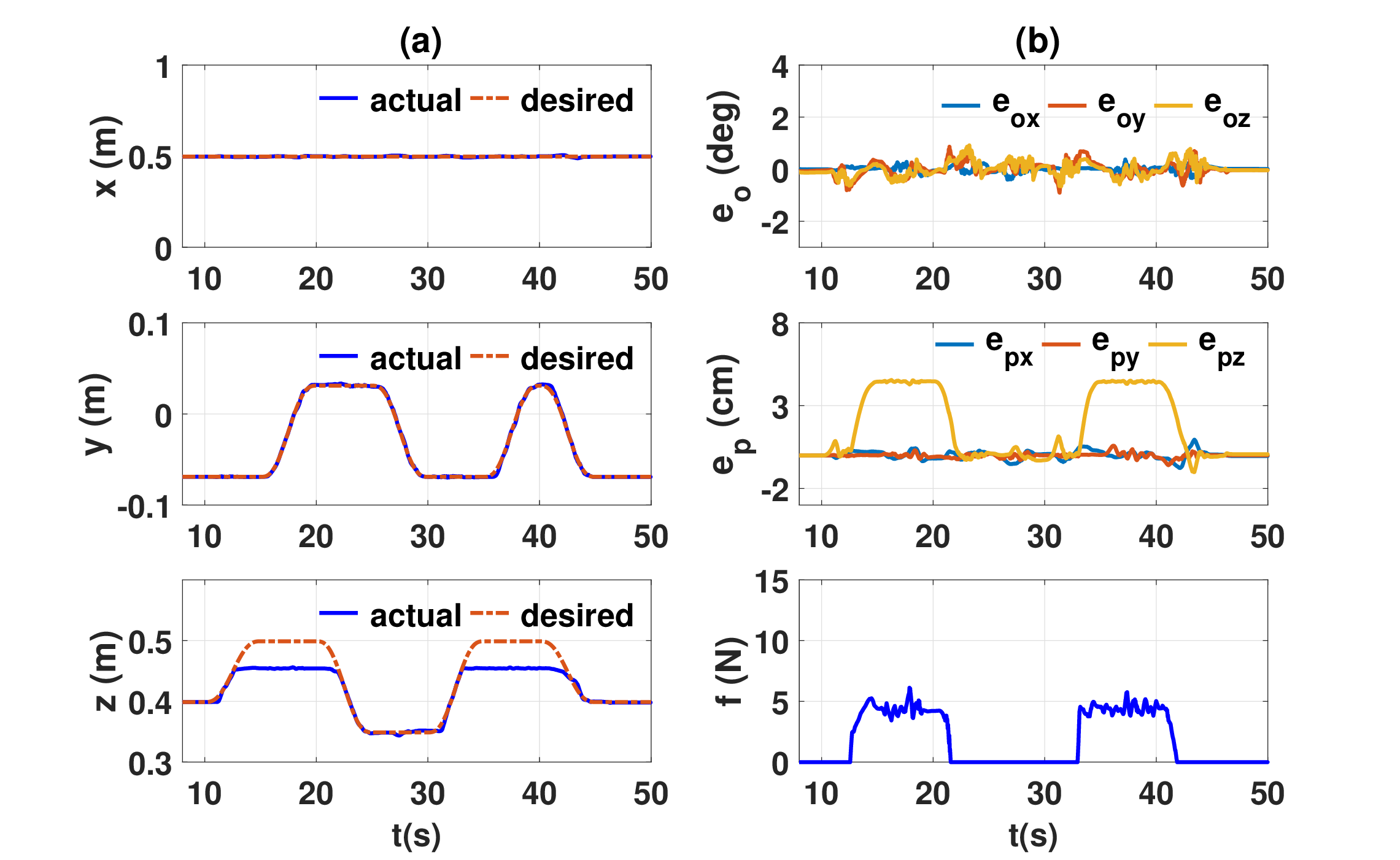}
      \caption{Experimental results of designed controller with \(k_e = 1000\, N/m\), \(m_d = 0.14\, kg\), and \(t_f = 5\, s\) with contact in \(z\) direction. a) position tracking, b) orientation error, position error, and contact force}
      \label{Exp5s1000}
   \end{figure}
   
\subsection{Experimental Results of Designed Controller}

The performance of the designed controller is examined in this section. The control goal of HULE is to track a desired square path in \(y-z\) plane with the desired orientation while carrying a human arm and having stable contact with a virtual wall placed in \(z\) direction. Two different paths with \(t_f =2s\) (fast trajectory) and \(t_f =5s\) (slow trajectory) are generated, and the performance of the controller is analyzed. The control parameters are selected as: \(\gamma = 10, K_{Di} = 0.5, K_{Ii} = 7, k_{di} = 0.05, k_{Ii} = 7\). The HULE robot is the ABLE robot \cite{c3}, manufactured by Haption. The control signals are transmitted back and forth to the robot using SIMULINK and Haption interface with a sample time of \(1\, ms\).

Fig. 5 demonstrates the performance of the designed control for the slow trajectory. Fig. 5(a) shows the tracking of desired trajectory in the \(x-y-z\) direction, while Fig. 5(b) shows the tracking errors of position and orientation along with the contact force. It can be seen that the orientation error is kept at less than \(1\, deg\). Moreover, the position tracking error in the \(x-y\) direction is less than \(1\, cm\), while stable contact with the environment is ensured in the \(z\) direction. Fig. 6 displays the results for the fast trajectory. It can be concluded from Fig. 6 that although the velocity of the end-effector at the contact point is increased, the contact is still stable, and the tracking error of position and orientation are less than \(1\, cm\) and \(2\, deg\), respectively. 
\begin{figure}[t]
    \centering
    \includegraphics[width=3in]{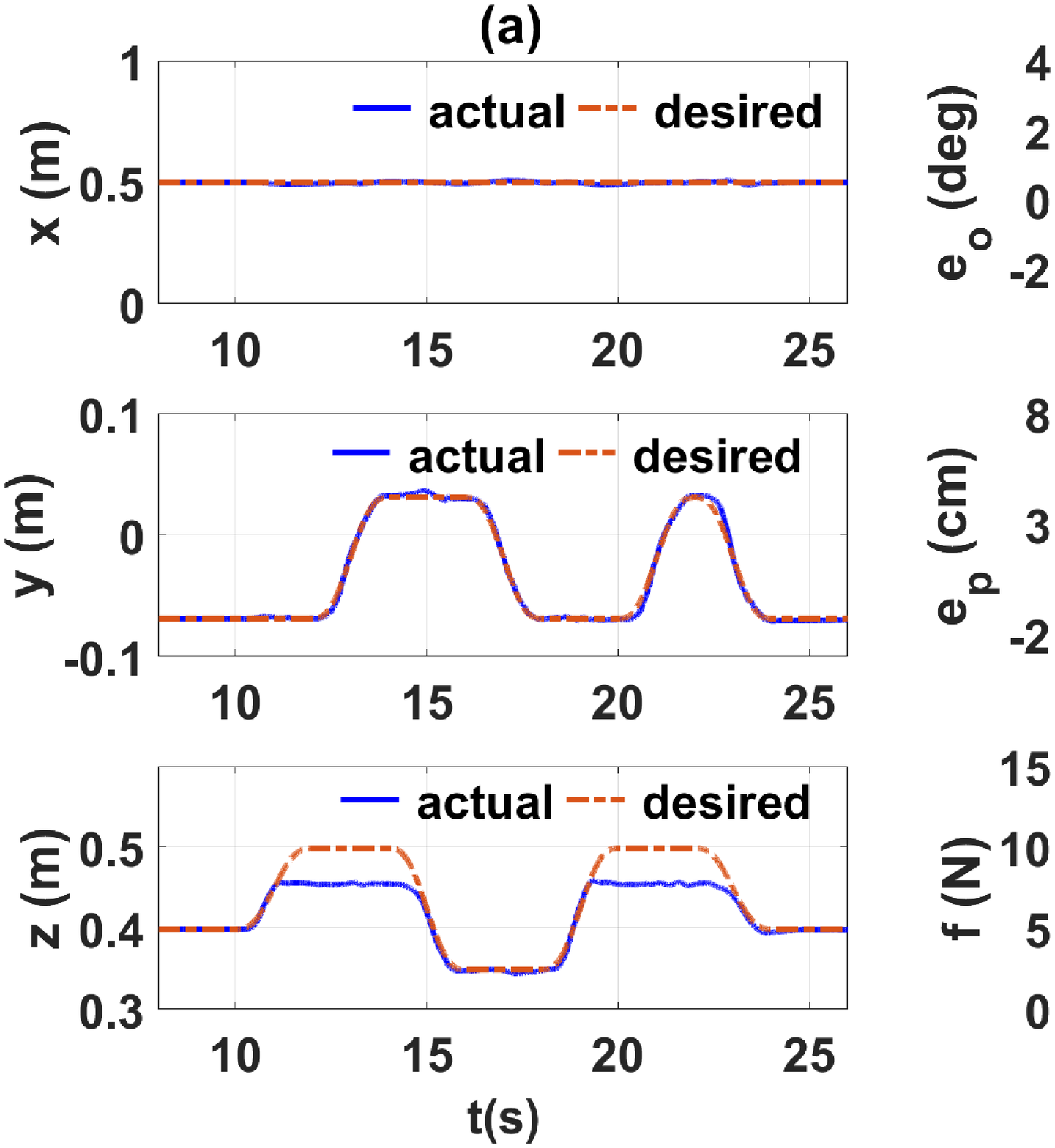}
      \caption{Experimental results of designed controller with \(k_e = 1000\, N/m\), \(m_d = 0.14\, kg\), and \(t_f = 2\, s\) with contact in \(z\) direction. a) position tracking, b) orientation error, position error, and contact force}
      \label{Exp2s1000}
   \end{figure}

   \begin{figure}[ht]
    \centering
    \includegraphics[width=3in]{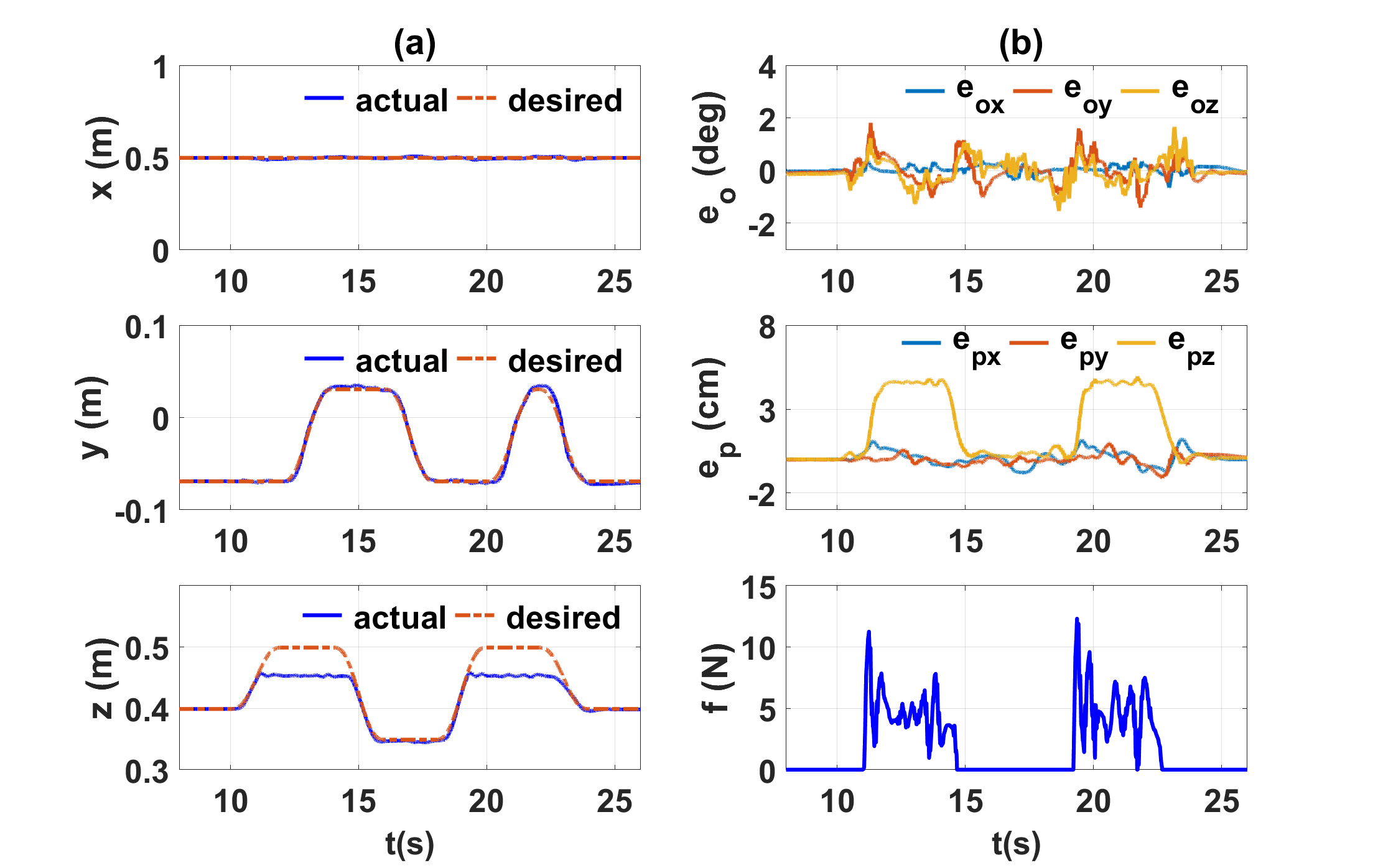}
      \caption{Experimental results of designed controller with \(k_e = 1500\, N/m\), \(m_d = 0.14\, kg\), and \(t_f = 2\, s\) with contact in \(z\) direction. a) position tracking, b) orientation error, position error, and contact force}
      \label{Exp2s1500}
   \end{figure}

To show the performance of the controller, the interaction with a virtual wall with \(k_e = 1500 \, N/m\) and \(t_f = 2\, s\) is examined in Fig. 7. The stiffness of the environment is increased along with the velocity of the contact, which can result in instability. It can be seen from Fig. 7 that not only is the contact stable, but the controller also shows good performance in keeping tracking errors considerably low. The performed analysis shows that the designed controller ensures the goals of pHREI in different scenarios. 

To have a better performance evaluation, the designed controller is compared to the state-of-the-art methods, such as neuro-adaptive backstepping impedance control (NABIC) \cite{c28} and model-free adaptive impedance (MFAI) control \cite{c29}. Control parameters are tuned to get the best performance for each controller. The control goal is the same as in Fig. 6. The root-mean-square (RMS) value of simulation results for position error in the z-direction (\(e_{p_z}\)) and xy-direction (\(e_{p_{xy}}\)) in \(mm\), orientation error \(e_o\) in \(deg\), and control action along with the maximum contact force are presented in Table I. It can be seen that the proposed controller demonstrated excellent performance in comparison to other methods for a given approximately the same control action. As explained in Remark 1, the perfect performance of the presented controller can be attributed to its subsystem-based approach that enables the controller to focus on the rigid body and actuator part separately.

\subsection{Discussion}
The difference between TDPA presented in \cite{c15} and VDC must be clarified. TDPA is a nonmodel-based approach that ensures the passivity of the system by adding virtual damping to the generic controller of the system, whereas VDC is a model-based approach that designs a controller to accomplish control objectives. Additionally, VDC does not need any virtual damping to stabilize the contact. The damping element utilized in this paper is only employed to expand the Z-width of the HULE, which shows the renderable impedance of the environment. In contrast to TDPA, in which the passivity term (passivity observer) is a crucial term to ensure the passivity, VDC can stabilize the contact with no need for damping elements for a limited region of Z-width. The expansion of the Z-width only helps to have a wider range of stability for VDC. Moreover, VPF in VDC may look like the passivity observer (PO) in TDPA in the way that they are defined. However, PO is utilized in TDPA to adjust control action, whereas VPF is only utilized in stability analysis and has no impact on control action.

\begin{table}[t]
\centering
\caption{Comparison of Control Methods}
\label{table}
\setlength{\tabcolsep}{10pt}
\begin{tabular}{|p{40pt}|p{15pt}|p{15pt}|p{15pt}|p{15pt}|p{15pt}|}
\hline
Controller& 
\(e_{p_{z}}\)& 
\(e_{p_{xy}}\)& 
\(e_o\)&
\(F_{max}\)&
\(tau\)
\\
\hline
MFAI \cite{c29} & 
29.21& 
3.5&
1.18&
6.55&
2.45\\
NABIC \cite{c28}&
29.88& 
3.65&
1.02&
4.7&
2.34\\
proposed&
23.39& 
1.49&
0.25&
5.02&
2.51\\
\hline
\end{tabular}
\label{tab}
\end{table}

\section{Conclusion}
In this study, a subsystem-based adaptive impedance control is designed for the pHREI control of a 7-DoF HULE. The VDC controller divided the complex system into subsystems and based on the VDC-impedance law (\ref{equ23}), the control law is designed in (\ref{equ14}) to achieve the desired impedance of (\ref{equ20}). The Z-width plot of the 7-DoF HULE was drawn, and it was shown that employing varying virtual mass can enhance the renderable impedances by the haptic display. Moreover, the performance of the designed controller was examined by performing some experiments and comparing to state-of-the-art control methods. Section VI shows that HULE with the presented controller demonstrates better performance and perfectly follows the desired path and tolerates human arm and virtual wall forces. 

\appendices
\section{Proof for Theorem 2}
\renewcommand{\theequation}{A.\arabic{equation}}
\setcounter{equation}{0}
Substituting (\ref{equ24}) and (\ref{equ20}) in (\ref{equ23}), we have,
\begin{equation}\label{A.1}
\begin{split}
        \Dot{X}_r & = \Dot{X}_d + \Gamma_x(X_d-X)-B_d^{-1}(B_d(\Dot{X}_d-\Dot{X})\\
        & + K_d(X_d-X)) = \Dot{X}_d + \Gamma_x(X_d-X) \\
        & -B_d^{-1}B_d(\Dot{X}_d-\Dot{X}) - B_d^{-1}K_d(X_d-X)\\
        & = 
        \Gamma_x(X_d-X)+\Dot{X}- \Gamma_x(X_d-X)= \Dot{X}
\end{split}
\end{equation}
Now, utilizing result of (\ref{A.1}) in (\ref{equ23}), one can obtain,
\begin{equation}\label{A.2}
\begin{split}
        f_d-f &=  -\Gamma_f^{-1}(\Dot{X}_d-\Dot{X}_r) - \Gamma_f^{-1} \Gamma_x(X_d-X)
        \\
        & = -B_d(\Dot{X}_d-\Dot{X})-K_d(X_d-X)
\end{split}
\end{equation}
which shows that the desired impedance in (\ref{equ20}) is achieved.

\section{Proof for Theorem 3}
\renewcommand{\theequation}{B.\arabic{equation}}
\setcounter{equation}{0}
Subtracting (\ref{equ11}) from (\ref{equ9}) and (\ref{equ15}) from (\ref{equ13}) along with using (\ref{equ12}) and (\ref{equ16}) result in,
\begin{equation}\label{B.1}
    \begin{split}
        M_{B_i}\frac{d}{dt}{\,^{B_i}e_{\mathcal{V}}}& =\, (^{B_i}F_r^*-\,^{B_i}F^*) - \, ^{B_i}W \Tilde{\phi}_{B_i} -C_{B_i}{\,^{B_i}e_{\mathcal{V}}}\\
        & - \,K_{Di}\,^{B_i}e_{\mathcal{V}} - \,K_{Ii}\,\int_{0}^{t} \,^{B_i}e_{\mathcal{V}} dt
    \end{split}
\end{equation}
\begin{equation}\label{B.2}
    \begin{split}
        I_{mi}\frac{d}{dt}{e_{ai}}& =\, (\tau_{ir}^*-\tau_i^*) - \, W_{ai} \Tilde{\phi}_{ai} - \,k_{di}\,e_{ai} - \,k_{Ii}\,\int_{0}^{t} \,e_{ai} dt
    \end{split}
\end{equation}
where \(\Tilde{\phi}_{B_i} = \hat{\phi}_{B_i}-\phi_{B_i}\) and \(\Tilde{\phi}_{ai} = \hat{\phi}_{ai}-\phi_{ai}\). Taking the time derivative of (\ref{equ25}), using Lemma 2, (\ref{B.1}), (\ref{B.2}), and conducting mathematical manipulation, we get to,
\begin{equation}\label{B.3}
    \begin{split}
        \Dot{\nu}_i (t) &= \sum_{i=1}^{7} [\,^{B_i}e_{\mathcal{V}}^T\,(^{B_i}F_r^*-^{B_i}F^*)-\,\Tilde{\phi}_{B_i}^T\,s_{B_i} \\
        &-\,^{B_i}e_{\mathcal{V}}^T\,K_{Di}\,^{B_i}e_{\mathcal{V}} +e_{ai}\,(\tau_{ir}^*-\tau_i^*)- \Tilde{\phi}_{ai}^T\,s_{ai}\\
        &-\,k_{di}e_{ai}^2+tr([\hat{\mathcal{L}}_{B_i}^{-1}\Dot{\hat{\mathcal{L}}}_{B_i}\,\hat{\mathcal{L}}_{B_i}^{-1}]\,\Tilde{\mathcal{L}}_{B_i})\\
        &+tr([\hat{\mathcal{L}}_{ai}^{-1}\Dot{\hat{\mathcal{L}}}_{ai}\,\hat{\mathcal{L}}_{ai}^{-1}]\,\Tilde{\mathcal{L}}_{ai})]
    \end{split}
\end{equation}
with \(s_{B_i} = \,W_{B_i}^T\,^{B_i}e_{\mathcal{V}}\) and \(s_{ai} = \,W_{ai}^T\,e_{ai}\). By rewriting (\ref{B.3}) using \(\Tilde{\phi}_{B_i}^T\,s_{B_i} = tr(\Tilde{\mathcal{L}}_{B_i} \mathcal{S}_{B_i})\) and \(\Tilde{\phi}_{ai}^T\,s_{ai} = tr(\Tilde{\mathcal{L}}_{ai} \mathcal{S}_{ai})\), and Lemma 1, we have 
\begin{equation}\label{B.4}
    \begin{split}
        \Dot{\nu}_i (t) &= \sum_{i=1}^{7} [\,^{B_i}e_{\mathcal{V}}^T\,(^{B_i}F_r^*-^{B_i}F^*)+e_{ai}\,(\tau_{ir}^*-\tau_i^*) \\
        &-\,^{B_i}e_{\mathcal{V}}^T\,K_{Di}\,^{B_i}e_{\mathcal{V}}-\,k_{di}e_{ai}^2].
    \end{split}
\end{equation}
The first two terms on the right side of (\ref{B.4}) are VPFs related to the rigid body and actuator parts. It can be shown that \cite{c23},
\begin{equation}\label{B.5}
    \begin{split}
        \,^{B_i}e_{\mathcal{V}}^T\,(^{B_i}F_r^*-^{B_i}F^*) = p_{B_i}-p_{T_i}
    \end{split}
\end{equation}
\begin{equation}\label{B.6}
    \begin{split}
        e_{ai}\,(\tau_{ir}^*-\tau_i^*) = -p_{B_i}+p_{T_{i-1}}
    \end{split}
\end{equation}
where \(p_{T_0}=0\) because the ground has no velocity. By replacing (\ref{B.5}) and (\ref{B.6}) in (\ref{B.4}), it can be concluded that all the VPFs cancel one another at the VCP, except that which represents contact with the environment,
\begin{equation}\label{B.7}
    \begin{split}
        \Dot{\nu}_i (t) = \sum_{i=1}^{7} {[-\,^{B_i}e_{\mathcal{V}}^T\,K_{Di}\,^{B_i}e_{\mathcal{V}}-\,k_{di}e_{ai}^2]-p_{T_7}.}
    \end{split}
\end{equation}
It is obvious from (\ref{B.7}) that if \(p_{T_7}\) vanishes, the asymptotic stability of the entire system is established. As the \({T_7}\) frame is attached to the end-effector of the robot, its velocity and force are velocity and force represented in Cartesian space. Therefore, utilizing the definition of VPF in (\ref{equ2}), \(^{T_7}V_r = \Dot{X}_r\), \(\,^{T_7}F_r = f_d\), (\ref{equ20}), (\ref{equ23}), and (\ref{equ24}) one can get,
\begin{equation}\label{B.8}
\begin{split}
    &p_{T_7}  = (\Dot{X}_d-\Dot{X})^T\,(B_d\,B_d^{-1}\,B_d-B_d)\,(\Dot{X}_d-\Dot{X})+\\
    & (\Dot{X}_d-\Dot{X})^T\,(2B_d^{-1}\,K_d\,B_d-K_d-B_d^{-1}\,K_d\,B_d)(X_d-X)\\
    & + (X_d-X)^T\,(K_d\,B_d^{-1}\,K_d-K_d\,B_d^{-1}\,K_d)\,(X_d-X)= 0.
\end{split}
\end{equation}
Finally, by replacing the result of (\ref{B.8}) in (\ref{B.7}), the stability of the entire system in the sense of Theorem 1 is ensured.

\vfill
\end{document}